# Blockchain associated machine learning and IoT based hypoglycemia detection system with auto-injection feature

**Rahnuma Mahzabin[1], Fahim Hossain Sifat[1], Sadia Anjum[1], Al-Akhir Nayan[2], Muhammad Golam Kibria[1]**

[1]Department of Computer Science and Engineering, School of Science and Engineering, University of Liberal Arts Bangladesh (ULAB), Dhaka, Bangladesh
[2]Department of Computer Engineering, Faculty of Engineering, Chulalongkorn University, Bangkok, Thailand

**ABSTRACT**

Hypoglycemia is an unpleasant phenomenon caused by low blood glucose. The disease can lead a person to death or a high level of body damage. To avoid significant damage, patients need sugar. The research aims at implementing an automatic system to detect hypoglycemia and perform automatic sugar injections to save a life. Receiving the benefits of the internet of things (IoT), the sensor's data was transferred using the hypertext transfer protocol (HTTP) protocol. To ensure the safety of health-related data, blockchain technology was utilized. The glucose sensor and smartwatch data were processed via Fog and sent to the cloud. A Random Forest algorithm was proposed and utilized to decide hypoglycemic events. When the hypoglycemic event was detected, the system sent a notification to the mobile application and auto-injection device to push the condensed sugar into the victim's body. XGBoost, k-nearest neighbors (KNN), support vector machine (SVM), and decision tree were implemented to compare the proposed model's performance. The random forest performed 0.942 testing accuracy, better than other models in detecting hypoglycemic events. The system's performance was measured in several conditions, and satisfactory results were achieved. The system can benefit hypoglycemia patients to survive this disease.

*Corresponding Author:*

Al-Akhir Nayan
Department of Computer Engineering, Faculty of Engineering, Chulalongkorn University
Bangkok, Thailand
Email: asquiren@gmail.com

## 1. INTRODUCTION

Hypoglycemia is a condition that shows symptoms of sweating, heart rate fluctuations, shaking, dizziness, and so on when blood glucose level goes under 70 mg/dl. The worst scenario can be seen when a patient gets senseless and sometimes dies [1]. This condition is often observed in diabetic patients who depend on insulin. So, when the patients get a hypoglycemic attack, they cannot get any kind of hint. They only realize when they feel weak or have any hypoglycemic symptoms, but they can't be sure without measuring the hypoglycemia. Here a third party is needed to measure the blood glucose level and give the patient some sugar [2]. In recent years, machine learning and IoT-based technologies have shown promising results in health [3]-[5], agriculture [6], [7], and other sectors [8] in making automated decisions and treatments.

IoT advanced technologies enable us to measure blood glucose levels and send those data over the network through the applications. A-state-of-art is that different auto injections have been devised to complete the insertion process automatically [9]. These auto injections are connected over a network to servers and can be instructed as needed. Again, these servers need to be protected. These servers hold health-related data which are valuable. While health-related data is transferred from the sensor to the fog and then sent to the cloud, blockchain adds security such as immutability, traceability, decentralization, and transparency, which gains the trust of any system user.

Previous research shows that hypoglycemia has a relation to blood glucose, blood pressure, heart rate, shivering, and sweating [10]. But researchers have ignored those parameters. Some have worked with only blood glucose, and others have worked with blood pressure. Smartwatches [11] and blood glucose sensors [12] provide these real-time parameters data. Adding fog, real-time data transmission latency can be reduced [13]. Different machine learning algorithms and modular approaches have been utilized to detect hypoglycemic events [14]. But a complete integrated system cannot be found or has not been proposed for hypoglycemia detection dedicatedly. So, this research attempts to detect hypoglycemia in real-time with all the previously ignored parameters. After detection, the auto-injection device injects the condensed sugar [15], [16] to save the patient from the hypoglycemic consequences. As these are health-related confidential data, providing security with blockchain technology is necessary [17]. Combining these technologies in real-time hypoglycemia detection with the features transmitted from sensors to the cloud makes a complete system that gains the user's trust and impacts the user's life. This research paves the way for an industrialized business.

The whole summary of this research is mentioned in the abstract. A general discussion about the topics is mentioned in the introduction. In section 2, the research gaps have been covered. In section 3, the proposed system is discussed briefly. Implementation and results are explained in section 4. Then the research is concluded with a conclusion.

## 2. RELATED WORKS

Plenty of research focuses on detecting hypoglycemic events using machine learning and IoT. Chow *et al.* [18] found a clear relationship between heart rate and hypoglycemia. Billard *et al.* [19] observed a relationship with the blood pressure in a hypoglycemic state. Their experiments detected hypoglycemia, but their system was manual for data collecting and detecting.

Maritsch *et al.* [20] collected data with a smartwatch and tried to establish a relationship with heart rate. Heart rate data were collected with an empatica E4 smartwatch, and glucose data was collected from a CGM. They achieved the best performance using gradient boosting decision tree.

Sudharsan *et al.* [21] researched hypoglycemia and their dataset contained self-monitored glucose values. Random forest, KNN, SVM, and Naïve Bayes were applied to locate exceptional accuracy. Those algorithms performed like human experts and predicted hypoglycemic events correctly.

Gia *et al.* [10] developed an IoT-based glucose monitoring system that uses real-time data received via networks. This technology effectively communicated with data and issued a push notice when abnormal blood glucose levels. This study paved the way for developing an Internet of Things-based glucose monitoring system with real-time data transfer and alerting capabilities.

Ruan *et al.* [22] used machine learning for hypoglycemia detection. Medical data were collected, and the analysis was performed using gradient boosting. The model performed well, giving 0.96 (AUROC) accuracy on clinically significant hypoglycemia.

Anawar *et al.* [13] found a solution to the time delay issue in data transportation with fog computing. They concluded that time delay issues arise when real-time data is preprocessed and sent to devices through cloud computing. They suggested for more rapid data transport infrastructure where information is swiftly stored and processed closer to data sources on local fog nodes.

Tanwar *et al.* [23] found blockchain technology is the most secure approach to keeping datasets safe in healthcare. It can alter health interactions, access to patient information, and device tracking, among other things. Its authentication technique confirms validity without knowing the user's identity, and only legitimate users can decrypt the information saved.

Previous research has ignored several essential parameters that directly influence hypoglycemic events. Those drawbacks paved the way for us to work on this research. In this study, we have tried to overcome previous issues by prioritizing essential parameters and boosting the accuracy of previous machine learning approaches in case of detecting hypoglycemia.

## 3. METHODS
### 3.1. Proposed system

Figure 1 represents the block diagram of the automatic hypoglycemia detection system associated with an intelligent medicine injecting device. Blockchain has been used to impart security to the user's confidential health information. In this system, the users will be notified of any hypoglycemic attack. The injecting device will automatically push the appropriate quantity of medicine, adjusted by the computer-based dosing controller using AP and ADS technology.

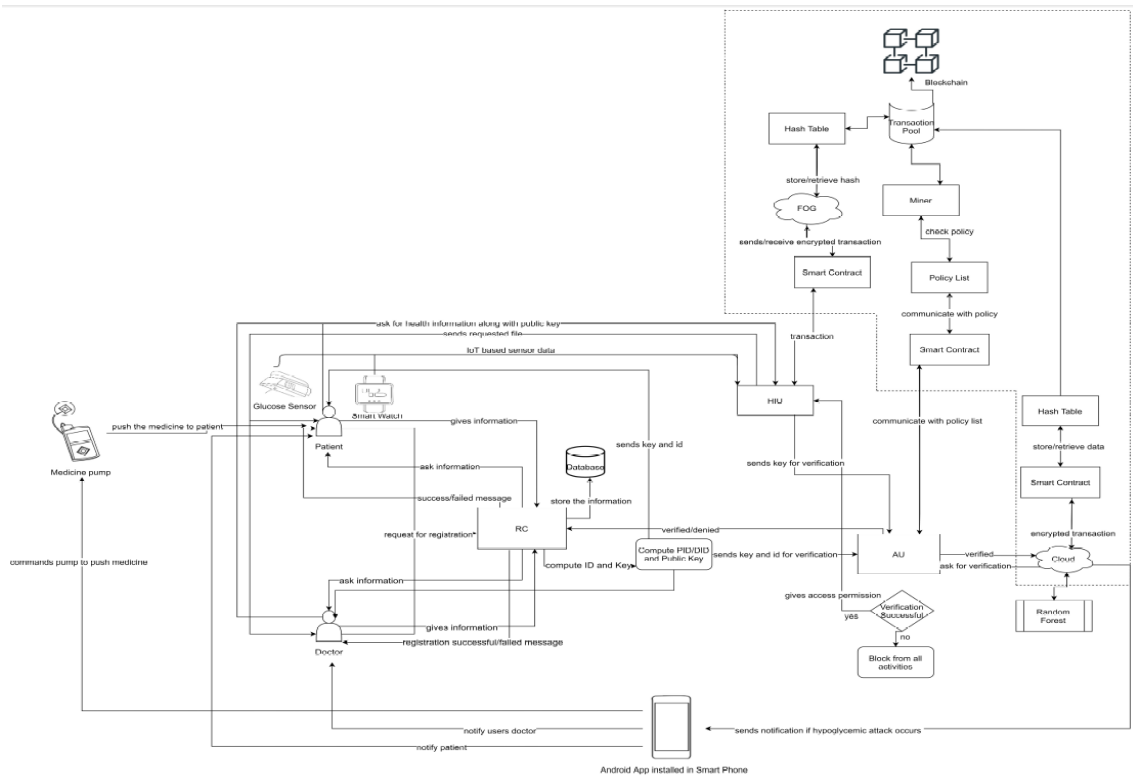

Figure 1. Proposed system block diagram

This system has two types of users. One is the patients who use this system, and another is the doctors who can only discern patients' data. Users need to be registered by transmitting a request to the registration center (RC). At the time of registration, a user needs to determine the user type, and the RC procures the relevant information by assigning an id and a public key. The Id and key are forwarded to the administration unit (AU) for further verification. The AU scrutinizes the user's authentication by the key and Id whenever they desire to interact with the system. Any unauthorized interaction blocks the user from all the activities of the system. Whenever any hypoglycemic attack occurs on any authorized user, the dosing controller optimizes the dose of that patient using AP and ADS and injects it into the user's body. A notification is sent to the user to apprise them about their health condition. All the health information of a particular user is stored through blockchain technology. The components of the system are described in detail.

The RC is used to register the users and computes the id and key of the users in this system. The RC procures information from the users and safely stores it in the database. It procures the name, date of birth, contact details (email, phone no, and address) from the patient and procures name, date of birth, contact details (email, phone no, and address), qualification, job details, and patient key and id from the doctors. It computes a unique id and public key for each user.

The IoT sensors and devices accumulate the patient's health information. The patients wear the sensors and devices, which are accumulated health information from the patient's body. The machine learning model uses health information to detect the hypoglycemic attack of the patients. In this system, patients need to wear a continuous glucose monitoring sensor to accumulate the glucose data of the patient's blood and a smartwatch to accumulate heart-rate data, systolic blood pressure, sweating, and shivering data from the patient body.

The health information unit (HIU) manages all the patient's health information. All the health data accumulated from a patient's body is placed in the blockchain through this unit. The patient's health data is proceeded to this unit along with the public key and id of that patient. This unit verifies the authenticity of the patient through the id and key with the aid of the AU. If the patient is authentic, the HIU proceeds the data accumulated from that patient to the blockchain. To place the data on the blockchain, the unit writes a smart contract for each transaction. If any patients or doctors desire to retrieve data from the blockchain, they must dispatch a request to the health information unit. The unit verifies the authenticity with the aid of the AU. If the requester acquires permission from the system, the HIU writes a smart contract for the transaction to communicate with the blockchain server. The requester acquires the requested data from the blockchain through the health information unit.

The AU governs all over the system and scrutinizes the user's authentication. At the time of registration, this unit investigates a user based on the id and public key that the registration center forwards. Moreover, it scrutinizes the authenticity of a user whenever the health information unit receives any health data from the patient, and any user desires to retrieve any data from the blockchain. The administration unit scrutinizes the authenticity based on the id and key of the user. It writes an intelligent contract to communicate with the policy list of the system. Following the strict policy list, the administration unit scrutinizes any user.

Fog computing makes the whole process of the system faster and securer. It processes the health data gathered from the patient's body using sensors and devices. The health data is moved to the Fog for being processed from the health information unit. Fog processed the data and forwarded them to the blockchain after processing. This system has used the cloud to operate the machine learning model of the system. After the machine learning model execution, the data the system receives is again placed on the blockchain from the cloud server.

The smart contract is the computer program written to control health data exchange in the system rightfully. The administration unit writes the intelligent contract to communicate with the policy list and the health information unit to exchange the health data. The hash table place the computed hashes of the transactions prepared to be placed on the blockchain. When the patient and the miners approve any transaction, it is considered to append to the blockchain network. The approved transaction hashes are stored in the hash table in this system. The transaction pool is where all the transactions are gathered, prepared to add to the blockchain, and retrieved from the blockchain. The health data and the detection result of a patient are gathered here to be appended to the blockchain. The requested data of a user is gathered here to be retrieved.

The machine learning model detects the hypoglycemic attack of the patients in this system. A patient's health data is retrieved from the blockchain in the cloud server. The trained model of the system is executed to detect the hypoglycemic attack using the retrieved data. The detection result is sent to the mobile system application to alert the user if an attack occurs. Five machine learning algorithms have been implemented for this system and compare the performance based on the accuracy. The implemented machine learning algorithms are KNN, gradient boosting classifier, random forest classification algorithm, SVM for kernel-linear, and SVM for kernel- RBF.

### 3.2. Proposed machine learning algorithm and training

Depending on the dataset's structural quality and the amount of data, an algorithm is required to perform better on an imbalanced dataset where overfitting issues can be observed. Besides, it also needs to handle null values or missing values so that those values cannot affect the outcome. Keeping those requirements in mind, the random forest classification algorithm was proposed and utilized for generating decisions about hypoglycemic events. The working dataset was split into test and train sets by a 20:80 ratio. K (K=5) fold cross-validation was addressed for splitting the dataset. Here sklearn ensembled random forest classifier was used. As the working dataset size was an average one, the number of trees was taken to 100, where each tree was given a max depth of 4 with 42 random states. After fitting into the model with hyperparameters tuning, it gave hypoglycemia detection decisions by averaging the tree's results. It reduced the number of false positives and increased the true positive results, which was the primary aim of this research.

### 3.3. Data accumulation and processing

Data were collected from several diabetic patients. 16,969 data was accumulated by the sensors and IoT devices from 16,969 diabetic patients. It took around 28 days to measure the data manually for 16,969 people. Total of seven features: glucose, diastolic blood pressure, systolic blood pressure, heart rate, body temperature, sweating, and shivering data are collected from each patient. The glucose data was gathered using a continuous glucose monitoring sensor named Dexcom CGM g6. The heart rate, diastolic blood pressure, systolic blood pressure, body temperature, and sweating data were accumulated by a smartwatch. The shivering data was collected manually. From these seven features, glucose, systolic blood pressure, heart rate, sweating, and shivering, these five features are related to hypoglycemic consequences. All the accumulated data was metamorphosed into float64 data types. The invalid data was replaced and handled by the mean value of the parameter.

The heart rate data was taken as a beat per second (bps) unit and metamorphosed into milliseconds (ms). The data processing time took around five days. After successful accumulation of data, according to the doctor's suggestion, hypoglycemia positive and negative data were classified. The working procedure of data preprocessing was done in a well-suited machine learning environment, google collaborator. A computer with a 64-bit Windows 10 operating system with an Intel Core i5 CPU was used to preprocess the data. For formalism and processing purposes, the python library was used. Table 1 represents a description of the sample dataset.

Table 1. Sample dataset description

|  | Glucose | Systolic blood pressure | Heart rate | Sweating | Shivering | Hypoglycemia |
|---|---|---|---|---|---|---|
| count | 16969.00 | 16969.00 | 16969.00 | 16969.00 | 16969.00 | 16969.00 |
| mean | 95.74 | 118.19 | 662.85 | 0.12 | 0.15 | 0.49 |
| std | 42.99 | 7.70 | 68.68 | 0.33 | 0.35 | 0.50 |
| min | 50.00 | 95.00 | 461.00 | 0.00 | 0.00 | 0.00 |
| 25% | 68.00 | 113.00 | 631.00 | 0.00 | 0.00 | 0.00 |
| 50% | 83.00 | 119.00 | 674.00 | 0.00 | 0.00 | 0.00 |
| 75% | 108.00 | 124.00 | 714.00 | 0.00 | 0.00 | 1.00 |
| max | 250.00 | 145.00 | 769.00 | 1.00 | 1.00 | 1.00 |

**3.4. Smartphone application**

A mobile application was implemented for the system to notify users about hypoglycemic attacks and medicine injections. The application can be installed by the users (patients and doctors). It instructs the medicine pump to inject medicine into the patient's body when a hypoglycemic attack occurs. The patients can view their hypoglycemic attack histories and current health data from the application. Moreover, the patient can view the quantity of medicine left in their medicine pump. It alerts the patients to refill the medicine before the last five doses.

**3.5. Auto injection and liquid sugar**

An automatic injection device is attached to the patient's body. Continuous subcutaneous insulin infusion (CSII) is used as a medicine pump for this system. The pump is connected to the system application. When any hypoglycemic attack occurs, the system application receives a notification from the cloud server. The mobile application sets the basal pump rate using ADS and AP. Then the application instructs the pump to insert the glucose dose (0.2 ml of glucose for adults and 0.1 ml of glucose for children) into the patient's body. After successfully inserting medicine, the pump sends a signal to the mobile app using Wi-Fi. After getting the signal, inform the user about the hypoglycemic attack and the current health information, quantity of pushed medicine, and the rest of the medicine. Then, if the notified person cannot reach the patient within 15 minutes, and after 15 minutes of the first dose of glucose, if the glucose level is still below 70mg/dl, it will again notify through the mobile application. The application instructs the pump to insert the glucose dose (0.2 ml of glucose for adults and 0.1 ml of glucose for children) into the patient's body. After giving the second dose of glucose, the pump again sends a signal to the mobile app using Wi-Fi. After getting the signal, it will again notify the user about the hypoglycemic attack and the current health information, quantity of pushed medicine, and rest of the medicine. Figure 2 represents the medicine pump.

The liquid-stable glucagon by Xeris pharmaceuticals is a pre-mixable solution so that it doesn't need manual mixing. The temperature will be 25 °C or 68 °F to 77 °F. The medicine's expiry date is 24 months after manufacture [24]. The auto-injector will be available in two pre-measured doses: a 0.5 mg/0.1 mL dose for pediatric administration and a one mg/0.2 mL dose for adolescent and adult administration [1], [2]. The liquid medicine must wrap in foil paper under the container as it needs to maintain 25 °C room temperature [25].

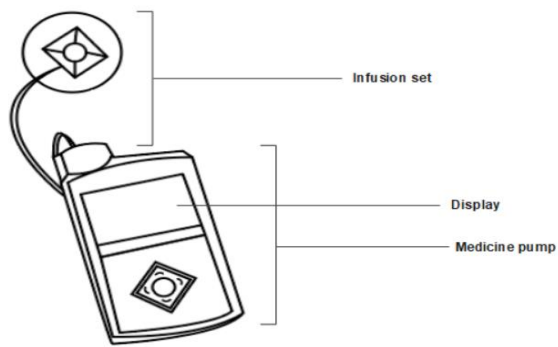

Figure 2. CSII medicine pump and automatic injecting device

### 3.6. Blockchain implementation

The blockchain was used to place health-related data securely. The data accumulated from the patient's body using the sensors and IoT devices is stored in the blockchain to keep it secure from unauthorized and third-party interaction. The detection result of the system's machine learning model is also appended to the blockchain with the approval of the miners. In this system, the patient, patient's doctor, and patient's relatives are considered miners. The authorized user can retrieve the health information or/and detection results from the blockchain with the approval of miners. Miner's approvals make the whole retrieving and appending procedure more secure and legitimate. In our system, when any health information is necessary to add to the blockchain, the miners are agreed upon to append the transaction to the blockchain and dispatch an agreement, including their signatures. The block's calculated hash is stored in the hash table. From the hash table, the transaction is assembled to the transaction pool and prepared to append to the blockchain. To more efficiently encrypt the blockchain data, the binary hash tree has been used where each leaf node is docketed with the cryptographic hash of a data block, and each non-leaf node is docketed with the cryptographic hash of its child nodes' dockets. Figure 3 represents the block structure of a block in the blockchain in this system. Some of the parameters are used to hash a block. The used parameters are- hash of the previous block, nonce, timestamp, Merkle tree, signatures of the miners, and the version of the block. In this research, the user id, hash, previous hash, index of the block, and root are set to 32bytes, and nonce and timestamp are set to 4bytes according to the privacy-preserving scheme of blockchain [23]. The Ethereum platform has been used to implement the blockchain network in this system. The SHA256 hash method has been used for hash computation that encrypts the data to a 256-bit signature.

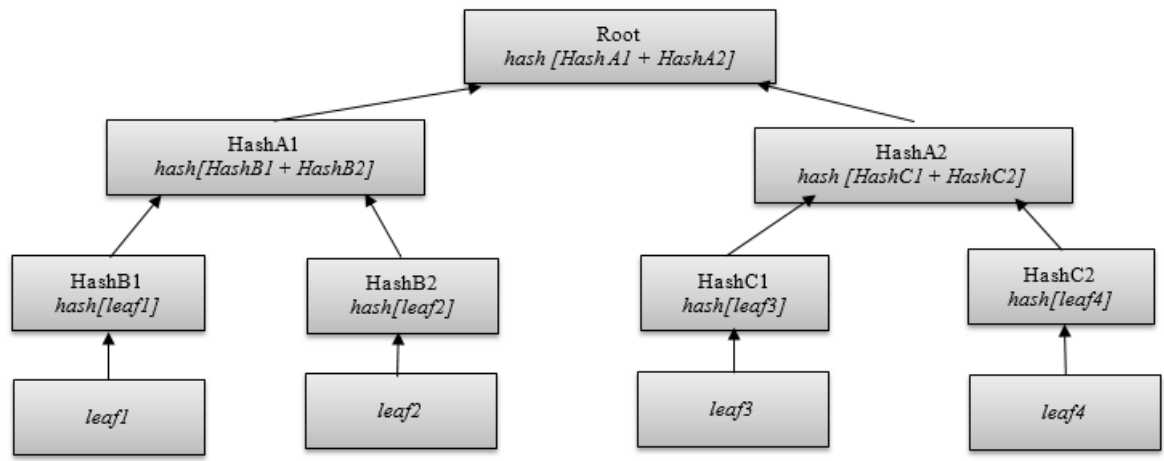

Figure 3. The block structure of a block in the blockchain

## 4. RESULT AND ANALYSIS
### 4.1. Performance comparison through machine learning

Five machine learning classification algorithms were used to detect hypoglycemic events and compare accuracy with the proposed method. The performances were examined based on the training and testing accuracy acquired by the implemented models. The machine learning classification algorithms were KNN, XGBoost, random forest, SVM (kernel-'RBF'), SVM ( kernel-'linear'), and decision tree. The models were implemented using the python library. Implementation and training of those models were performed using the google Colab environment. A 64-bit Windows 10 operating system with an Intel Core i5 CPU was used for writing code and uploading it to Google Colab.

Figure 4 represents the training and testing accuracies acquired from several classification algorithms. The random forest (RF) classification algorithm acquires 0.957 training and 0.942 testing accuracy. The random forest classifier algorithm was tuned with the hyperparameters' value of 'bootstrap': [True], 'max_depth': [3, 6, 10], 'max_features': [2, 3, 5, 7, 9], 'min_samples_leaf': [3, 4, 5, 9, 12], 'min_samples_split': [8, 10, 12] and 'n_estimators': [100, 200, 300, 500]. The gradient boosting algorithm was trained with 'max_depth': [3,6,10], 'learning_rate': [0.01, 0.05, 0.1], 'n_estimators': [100, 500, 1000] and 'colsample_bytree': [0.3, 0.7] hyperparameters' value and acquired 0.949 of training accuracy and 0.938 of testing accuracy. The KNN algorithm was implemented for seven different neighbors' values. For K= 9 and 11, the maximum accuracies were achieved. SVM algorithm was implemented for kernel values of 'linear' and 'RBF.' For both instances, the hyperparameters 'C' and 'gamma' were set to 1.0. For SVM kernel='RBF,' the acquired accuracy was 0.945 for training and 0.932 for testing. And for the kernel= 'linear' acquired 0.733 of accuracy for training and 0.721 for testing. The decision tree (DT) algorithm was implemented and obtained 0.948 training accuracy and 0.935 testing accuracy.

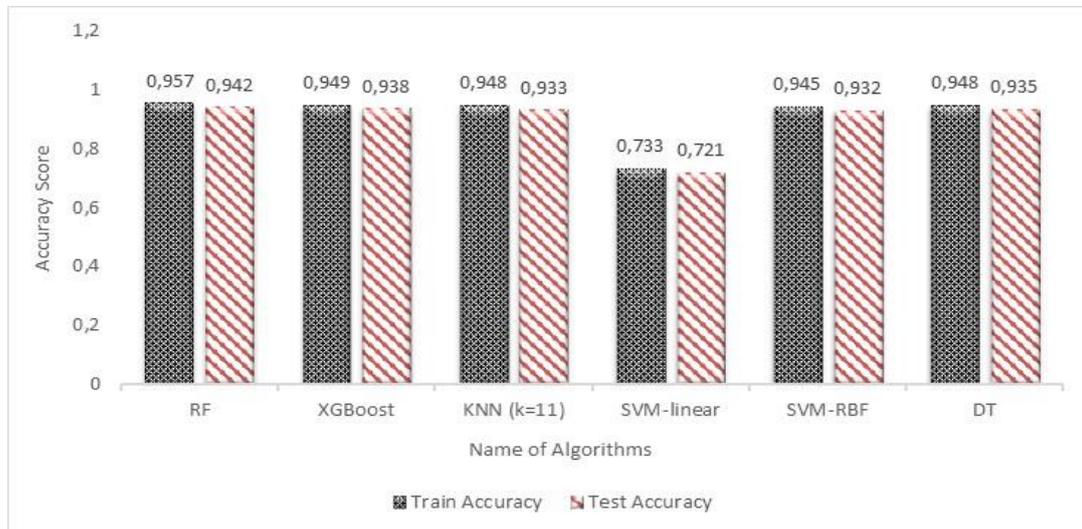

Figure 4. The accuracy comparison

Comparing the acquired accuracy of the implemented model, we found that the random forest (RF) algorithm provided supreme training accuracy. It performed excellently based on training and testing accuracy. The gradient boosting algorithm also imparted better accuracies close to RF, and the XGBoost occupied the second leading position. The DT algorithms occupied the third position. It acquired better testing accuracy than KNN, SVM-RBF, and SVM-linear. With 0.948 of training and 0.933 of testing accuracy, KNN (k=11) algorithm occupied the fourth position. The SVM-RBF classification algorithm occupied the fifth position. The SVM (kernel-'linear') algorithm achieved the lattermost position.

Figure 5 illustrates the ROC curve of the implemented machine learning models. The ROC curve demonstrates the performance of a particular model. It is a false positive rate vs. a true positive rate curve illustrating the performance of a model according to the actual positive and false-positive rates for multiple threshold values. The higher the ROC curve is, the better the model performs. The ROC curve of the RF model is higher than the other models. Hence the RF performance is more standard and satisfactory than other models.

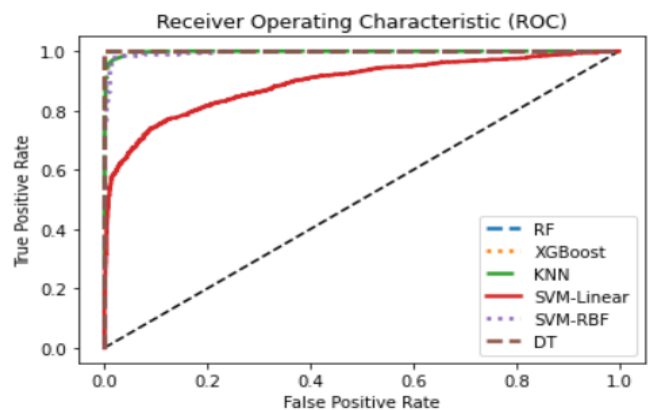

Figure 5. ROC curve of examined classification algorithms

## 5. CONCLUSION

Hypoglycemia is dangerous that can lead a person to death without providing any noticeable symptoms. In this research, we have tried to offer an automatic device to care for a person infected by the disease. A successful data transfer from the sensors to mobile applications was done, and the blockchain ensured data security. Different machine learning models were implemented for analyzing and detecting hypoglycemic events where the random forest algorithm beat the other four models in terms of accuracy. Finding the accurate model for the dataset was a tricky task. Adjusting auto-injection was challenging, but the device performed well in several test cases. It can be concluded that the device can take care of a hypoglycemia-infected person. We plan to improve the hypoglycemia detecting algorithm for faster and more accurate performance in the future.

## ACKNOWLEDGEMENT

The IoT Lab has supported this research work, a state-of-art specialized research facility of its kind situated at ULAB, implemented and supported by the Bangladesh Hi-Tech Park Authority of ICT Division.

## REFERENCES

[1] B. P. Childs, N. G. Clark, D. J. Cox, and P. E. Cryer, "Defining and reporting hypoglycemia in diabetes: a report from the American Diabetes Association Workgroup on Hypoglycemia," *Diabetes Care,* vol. 28, no. 5, pp. 1245–1249, 2005, doi: 10.2337/diacare.28.5.1245.
[2] P. E. Cryer, "Mechanisms of hypoglycemia-associated autonomic failure and its component syndromes in diabetes," *Diabetes*, vol. 54, no. 12, pp. 3592–3601, 2005, doi: 10.2337/diabetes.54.12.3592.
[3] A. Dey, K. A. Haque, A. -A. Nayan and M. G. Kibria, "IoT based smart inhaler for context-aware service provisioning," in *2020 2nd International Conference on Advanced Information and Communication Technology (ICAICT),* 2020, pp. 410-415, doi: 10.1109/ICAICT51780.2020.9333427.
[4] M. O. Rahman *et al.,* "Internet of things (IoT) based ECG system for rural health care*," Int. J. Adv. Comput. Sci. Appl.*, vol. 12, no. 6, pp. 470-477, 2021, doi: 10.14569/ijacsa.2021.0120653.
[5] A.-A. Nayan, B. Kijsirikul, Y. Iwahori, "Coronavirus disease situation analysis and prediction using machine learning: a study on Bangladeshi population," *International Journal of Electrical and Computer Engineering (IJECE),* vol. 12, no. 4, pp. 4217-4227, 2022, doi: 10.11591/ijece.v12i4.pp4217-4227.
[6] M. M. Rashid, A.-A. Nayan, S. A. Simi, J. Saha, M. O. Rahman, and M. G. Kibria, "IoT based smart water quality prediction for biofloc aquaculture," *Int. J. Adv. Comput. Sci. Appl.*, vol. 12, no. 6, pp. 56-62, 2021, doi: 10.14569/IJACSA.2021.0120608.
[7] A.-A. Nayan, J. Saha, A. N. Mozumder, K. R. Mahmud, A. K. Al Azad, and M. G. Kibria, "A machine learning approach for early detection of fish diseases by analyzing water quality," *Trends Sci*, vol. 18, no. 21, pp. 351, 2021, doi: 10.48048/tis.2021.351.
[8] A. -A. Nayan, M. G. Kibria, M. O. Rahman and J. Saha, "River water quality analysis and prediction using GBM," in *2020 2nd International Conference on Advanced Information and Communication Technology (ICAICT)*, 2020, pp. 219-224, doi: 10.1109/ICAICT51780.2020.9333492.
[9] D. Brand-Eubanks, "Gvoke HypoPen: An auto-injector containing an innovative, liquid-stable glucagon formulation for use in severe acute hypoglycemia," *Clin. Diabetes*, vol. 37, no. 4, pp. 393–394, 2019, doi: 10.2337/cd19-0040.
[10] T. N. Gia *et al.,* "IoT-based continuous glucose monitoring system: A feasibility study," *Procedia Comput. Sci.*, vol. 109, pp. 327–334, 2017, doi: 10.1016/j.procs.2017.05.359.
[11] M. Kheirkhahan *et al.,* "A smartwatch-based framework for real-time and online assessment and mobility monitoring," *J. Biomed. Inform.,* vol. 89, pp. 29–40, 2019, doi: 10.1016/j.jbi.2018.11.003.
[12] A. Hayek, A. A. Robert, and A. A. Dawish, "Evaluation of freestyle libre flash glucose monitoring system on glycemic control, health-related quality of life, and fear of hypoglycemia in patients with Type 1 Diabetes," *Clin. Med. Insights Endocrinol. Diabetes*, vol. 10, pp. 1–6, 2017, doi: 10.1177/1179551417746957.
[13] M. R. Anawar *et al.,* "Fog computing: An overview of big IoT data analytics," *Wirel. Commun. Mob. Comput.,* vol. 2018, pp 1–22, 2018, doi: 10.1155/2018/7157192.
[14] S. L. Cichosz, J. Frystyk, O. K. Hejlesen, L. Tarnow, and J. Fleischer, "A novel algorithm for prediction and detection of hypoglycemia based on continuous glucose monitoring and heart rate variability in patients with type 1 diabetes*," J. Diabetes Sci. Technol.*, vol. 8, no. 4, pp. 731–737, 2014, doi: 10.1177/1932296814528838.
[15] V. Valentine, B. Newswanger, S. Prestrelski, A. D. Andre, and M. Garibaldi, "Human factors usability and validation studies of a glucagon autoinjector in a simulated severe hypoglycemia rescue situation," *Diabetes Technol.* Ther., vol. 21, no. 9, pp. 522–530, 2019, doi: 10.1089/dia.2019.0148.
[16] R. M. Bergenstal *et al.,* "Safety of a hybrid closed-loop insulin delivery system in patients with type 1 diabetes," *JAMA*, vol. 316, no. 13, pp. 1407, 2016, doi: 10.1001/jama.2016.11708.
[17] G. P. Forlenza, "Use of artificial intelligence to improve diabetes outcomes in patients using multiple daily injections therapy," *Diabetes Technol. Ther.,* vol. 21, no. S2, pp. S24–S28, 2019, doi: 10.1089/dia.2019.0077.
[18] E. Chow *et al.,* "Risk of cardiac arrhythmias during hypoglycemia in patients with type 2 diabetes and cardiovascular risk," *Diabetes*, vol. 63, no. 5, pp. 1738–1747, 2014, doi: 10.2337/db13-0468.
[19] S. Feldman-Billard, P. Massin, T. Meas, P.-J. Guillausseau, and E. Héron, "Hypoglycemia-induced blood pressure elevation in patients with diabetes," *Arch. Intern. Med.*, vol. 170, no. 9, pp. 829–831, 2010, doi:10.1001/archinternmed.2010.98
[20] M. Maritsch *et al.,* "Towards wearable-based hypoglycemia detection and warning in diabetes," in *Extended Abstracts of the 2020 CHI Conference on Human Factors in Computing Systems,* Apr. 2020, pp. 1–8, doi: 10.1145/3334480.3382808.
[21] B. Sudharsan, M. Peeples, and M. Shomali, "Hypoglycemia prediction using machine learning models for patients with type 2 diabetes," *J. Diabetes Sci. Technol.*, vol. 9, no. 1, pp. 86–90, 2015, doi: 10.1177/1932296814554260.
[22] Y. Ruan *et al.,* "Predicting the risk of inpatient hypoglycemia with machine learning using electronic health records," *Diabetes Care,* vol. 43, no. 7, pp. 1504–1511, 2020, doi: 10.2337/dc19-1743.

[23] S. Tanwar, K. Parekh, and R. Evans, "Blockchain-based electronic healthcare record system for healthcare 4.0 applications," *J. Inf. Secur. Appl.,* vol. 50, no. 102407, 2020.
[24] W. Lane, "A renaissance in glucagon: Ready to use, ready to have!," *Taking Control Of Your Diabetes,* Aug. 2020.
[25] N. K. Giang, M. Blackstock, R. Lea, and V. C. M. Leung, "Developing IoT applications in the Fog: A distributed dataflow approach," *2015 5th International Conference on the Internet of Things (IOT),* 2015, pp. 155-162, doi: 10.1109/IOT.2015.7356560.

## BIOGRAPHIES OF AUTHORS

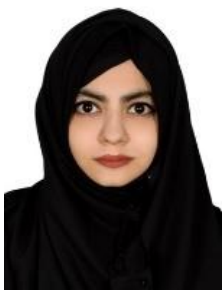

**Rahnuma Mahzabin** received Bachelor of Science degree in Computer Science and Engineering from University of Liberal Arts Bangladesh (ULAB), Dhaka, Bangladesh. She does research in Machine Learning and IoT. She can be contacted through the e-mail: rahnuma.mahzabin.cse@ulab.edu.bd.

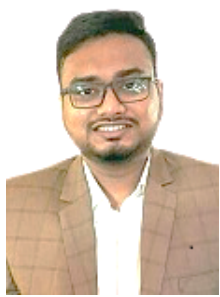

**Fahim Hossain Sifat** received Bachelor of Science degree in Computer Science and Engineering from University of Liberal Arts Bangladesh (ULAB), Dhaka, Bangladesh. He does research in Machine Learning and IoT. He can be contacted through the e-mail: fahimsifat29@gmail.com.

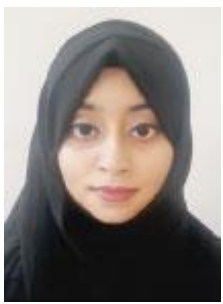

**Sadia Anjum** received Bachelor of Science degree in Computer Science and Engineering from University of Liberal Arts Bangladesh (ULAB), Dhaka, Bangladesh. She does research in Machine Learning and IoT. She can be contacted through the e-mail: sadia.anjum1.cse@ulab.edu.bd.

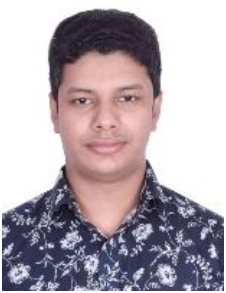

**Al-Akhir Nayan** received Bachelor of Science degree in Computer Science and Engineering from University of Liberal Arts Bangladesh (ULAB), Dhaka, Bangladesh. He is enrolling in a master's program at Chulalongkorn University, Bangkok, Thailand. He does research in Machine Learning, Artificial Neural Networks, and IoT. He can be contacted at email: asquiren@gmail.com.

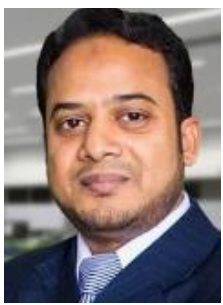

**Muhammad Golam Kibria** received his Ph.D. degree from the Department of Computer and Information Communications Engineering (CICE) at Hankuk University of Foreign Studies in Korea. His research interest includes the Internet of Things (IoT), Semantic Web, Ontology, and Web of Objects (WoO). He can be contacted at email: golam.kibria@ulab.edu.bd.